\documentclass{article}

\usepackage{times}
\usepackage{graphicx} 
\usepackage{subfigure} 

\usepackage{natbib}

\usepackage{algorithm}
\usepackage{algorithmic}

\usepackage{hyperref}

\usepackage{url}
\usepackage{amsmath,amssymb,amsthm}
\usepackage{multirow} 
\usepackage{picins}
\usepackage{graphicx} 
\usepackage[all]{xy}

\newcommand{\R}{\mathcal{R}}

\newcommand{\X}{\mathcal{X}}
\newcommand{\Y}{\mathcal{Y}}
\newcommand{\Z}{\mathcal{Z}}

\usepackage[accepted]{icml2015_arxiv}

\icmltitlerunning{Probabilistic Zero-shot Classification with Semantic Rankings}

\begin{document} 

\twocolumn[
\icmltitle{Probabilistic Zero-shot Classification with Semantic Rankings}
\icmlauthor{Jihun Hamm}{hammj@cse.ohio-state.edu}
\icmladdress{Computer Science and Engineering, The Ohio State University, Columbus, OH 43210, USA}
\icmlauthor{Mikhail Belkin}{mbelkin@cse.ohio-state.edu}
\icmladdress{Computer Science and Engineering, The Ohio State University, Columbus, OH 43210, USA}


\vskip 0.3in
]

\begin{abstract} 
In this paper we propose a non-metric ranking-based representation of semantic similarity
that allows natural aggregation of semantic information from multiple heterogeneous sources. 
We apply the ranking-based representation to zero-shot learning problems, and
present deterministic and probabilistic zero-shot classifiers which can
be built from pre-trained classifiers without retraining.
We demonstrate their the advantages on two large real-world image datasets. 
In particular, we show that aggregating different sources of semantic information,
including crowd-sourcing, leads to more accurate classification. 
\end{abstract} 

\section{Introduction}

\begin{figure*}[t]
\begin{center}
\includegraphics[width=1\linewidth]{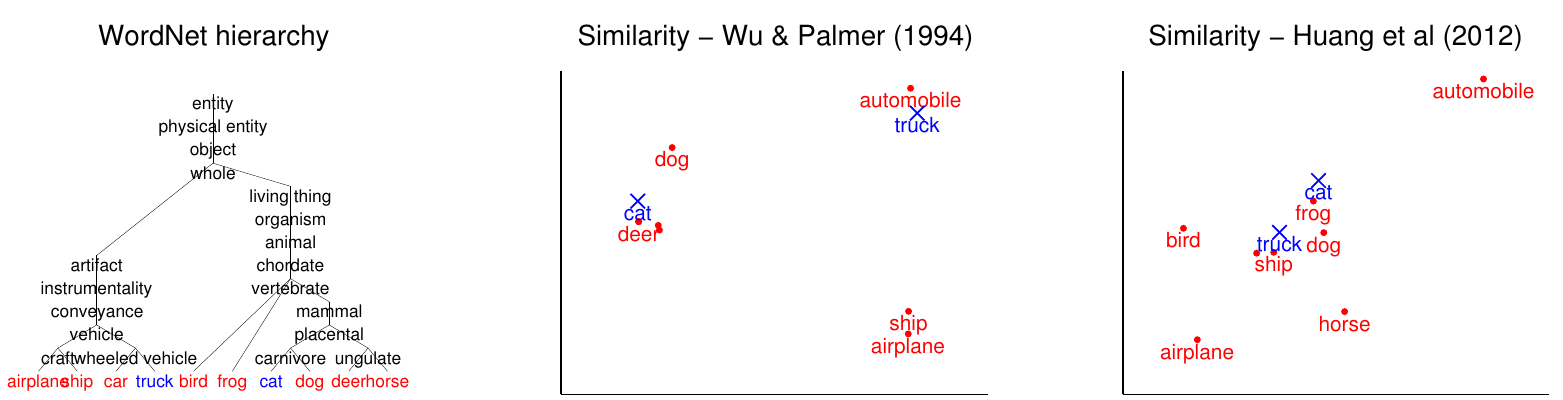}
\end{center}
\caption{
Representations of semantic relatedness of 10 objects.
Left: hierarchical tree from WordNet.
Middle: MDS embedding using Wu \& Palmer metric. 
Right: MDS embedding from \cite{Huang:2012}.
Note that the two embeddings from different similarity measures look very different
(Middle vs Right).
Some labels are hidden to avoid clutter.
}
\label{fig:10classes}
\end{figure*}

In standard multiclass classification settings, 
classes are treated as a categorical set without any extra structure. 
When we have side-information on the structure of classes, such as
semantic relatedness, we can use this information to improve 
the classification itself, or transfer any knowledge learned from the training domain
to solve problems in a new domain.

Consider a classification problem of the following 10 visual objects: 
{\it airplane, automobile, bird, cat, deer, dog, frog, horse, ship,} and {\it truck}.
There are many sources from which semantic information for those objects can be
obtained.
WordNet is a knowledge-base of semantic hierarchies developed manually
by linguistic experts \cite{Miller:1995}. 
In WordNet, objects form a hierarchical tree (Figure~\ref{fig:10classes}, Left),
where a child object is `a kind of' its parent object. 
Several similarity metrics can be defined from the hierarchy
\footnote{\url{http://maraca.d.umn.edu/similarity/measures.html}},
one of which is shown in Figure~\ref{fig:10classes} (Middle) as a two-dimensional
classical multidimensional scaling (MDS) embedding.
Semantic relatedness can also be mined automatically from existing corpora, 
such as Wikipedia, Google N-Gram corpus, or using search engines, where
cosine angles of co-occurrence vectors can be used as a similarity of two words.
More recently, elaborate methods for learning vectorial representations of words
have also been proposed \cite{Huang:2012,Mikolov:2013,Pennington:2014}.
Figure~\ref{fig:10classes} (Right) is an example MDS embedding from the representation 
from \cite{Huang:2012}.
As can be seen from the figure, similarity of the same objects can
look very different depending on which semantic source and measure is used.

{\bf Non-metric representation of similarity}. 
Multiple sources of semantic information have the potential to complement each other
for an improved classification result.
Still, how to best aggregate similarity from inhomogeneous sources remains an open problem. 
Similarity measures from different corpora or methods are not directly comparable,
and therefore a simple averaging of the measures will not be optimal.
The first key idea of our paper is that we use {\it non-metric, ranking-based}
representation of semantic similarity, instead of numerical representation.

To illustrate our approach, consider the problem of distinguishing {\it cat} and {\it truck}.
In Figure~\ref{fig:10classes} (Middle), {\it cat} is closer to {\it dog} than 
{\it automobile}:
\[
d({\it cat}, {\it dog}) < d({\it cat}, {\it automobile}),
\]
and {\it truck} is closer to {\it automobile} than to {\it dog}:
\[
d({\it truck}, {\it automobile}) < d({\it truck}, {\it dog})).
\]
In other words, we may be able to distinguish {\it cat} and {\it truck} from their 
closeness to other reference objects without using any numerical similarity. 
As a special case, we can use the similarity of all the other objects to {\it cat}
to form a {\it semantic ranking} of {\it cat}. 
For example, {\it cat} has a semantic ranking
\begin{equation}\label{eq:ranking of cat}
\pi_{\mathrm{\it cat}} = [{\it horse}, {\it deer}, \cdots, {\it automobile}, {\it airplane}],
\end{equation}
and {\it truck} has a semantic ranking
\begin{equation}\label{eq:ranking of truck}
\pi_{\mathrm{\it truck}} = [{\it automobile},{\it ship}, \cdots, {\it deer}, {\it horse}],
\end{equation}
according to the distance in Figure~\ref{fig:10classes} (Middle).
Not only the ordinal similarity may be sufficient for distinguishing {\it cat}
and {\it truck}, but it also seems a more natural representation, 
since the ordinal similarity is invariant under scaling and monotonic 
transformation of numerical values and therefore has a better chance of
being consistent across different heterogeneous sources.
Moreover, ordinal information can be obtained directly from non-numerical comparisons.
In particular, when we ask human subjects to judge similarity of objects, 
it is easier for subjects to rank objects rather than to assign
numerical scores of similarity. 

{\bf Zero-shot classification without retraining}. 
In this paper, we apply non-metric rankings-based representations of semantic similarity
to zero-shot classification problems~\cite{Palatucci:2009:NIPS,Lampert:2009:CVPR,Rohrbach:2010:CVPR,Rohrbach:2011:CVPR,
Qi:2011:CVPR,Mensink:2012:ECCV,Frome:2013:NIPS,Socher:2013:NIPS}. 
In zero-shot learning 
we have samples $\{(x_i, y_i)\}$ from the domain $\X \times \Y$
(e.g., $\Y$ is the set of 8 objects), 
but no samples from the test domain $\X \times \Z$ (e.g., $\Z=\{{\it cat},{\it truck}\})$.
The goal is to construct a classifier $\X \to \Z$ using the only training data 
$\{(x_i, y_i)\}$ and semantic knowledge of the two domains $\Y$ and $\Z$.

\begin{figure}[htb]
\begin{center}
\includegraphics[width=0.65\linewidth]{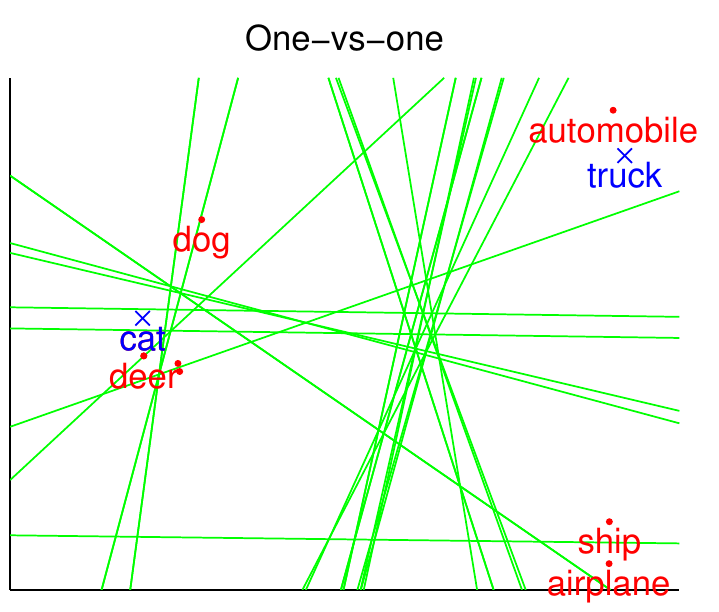}
\end{center}
\caption{
Decision hyperplanes for classifying 8 objects partition the feature space into
cells that correspond to rankings (see text). The lines are $\frac{8\cdot7}{2}$
separating hyperplanes from the one-vs-one binary classification setting.
}
\label{fig:one-vs-one}
\end{figure}

A standard approach to classifying $C$ classes is to use binary
classifications in one-vs-rest or one-vs-one setting, or to use multiclass losses directly.
If we already have pre-trained classifiers of the training domain classes $\Y$
using one of those settings, 
can we use those classifiers `for free' to distinguish unseen classes {\it cat}
and {\it truck} without re-training with training domain samples?
Figure~\ref{fig:one-vs-one} provides an intuition on the problem.
Consider multiple decision hyperplanes learned from the one-vs-one setting
(others will be discussed in Section~\ref{sec:zero-shot learning}.)
The $\frac{C(C\mathrm{-}1)}{2}$ hyperplanes partition the feature space into `cells', 
each of which assigns a {\it ranking} of $C$ objects to points inside its interior.
To see this, note that all pairs of objects are compared in each cell 
(either $i \prec j$ or $j \prec i$), and transitivity 
(see Section~\ref{sec:zero-shot learning})
follows the metric triangle inequality.
The ranking of an unseen test sample assigned by pre-trained classifiers
can be compared with the semantic rankings of {\it cat} or {\it truck} for
zero-shot classification, assuming feature and semantic similarities are strongly
correlated (see \cite{Deselaers:2011:CVPR} for a discussion).

\vspace{-0.2in}
Building on this idea, we present novel zero-shot classification methods that 
are free of re-training and can aggregate semantic information from multiple sources.   
We start by proposing a simple deterministic ranking-based method, and further
improve the method by introducing probability models of rankings. 
In the probabilistic approach, real-valued classification scores
are mapped to posterior probabilities of rankings, and combined with prior probability
of rankings learned from (multiple) semantic sources. 
The advantage of using probabilistic approach will be explained more in the method and
the experiment sections. 
For both the posterior and the prior probabilities of rankings,
we use classic probabilistic models of ranking including the Plackett-Luce, the Mallows, 
and the Babington-Smith models.

To summarize the contributions of this paper, we present
\begin{enumerate}\setlength{\itemsep}{-3pt}
\item non-metric ranking-based representation of a semantic structure,
alternative to numerical similarity representation
\item methods of aggregating multiple semantic sources using probability models of rankings
\item deterministic and probabilistic zero-shot classifiers built from pre-trained classifiers
without retraining.
\end{enumerate}
\vspace{-0.1in}
In the experiment section we demonstrate the advantages of 
our approach over a numerically-based approach and a deterministic approach
using two well-known image databases
Animals-with-attributes \cite{Lampert:2009:CVPR} and CIFAR-10/100 \cite{Krizhevsky:2009:TR}. 
In particular, we demonstrate that aggregating different semantic sources,
including crowd-sourcing, leads to more accurate zero-shot classification. 

The remainder of the paper is organized as follows: 
In Section~\ref{sec:zero-shot learning}, we present deterministic 
and probabilistic ranking-based algorithms for zero-shot classification.
In Section~\ref{sec:related work}, we relate our work to others in the literature.
In Section~\ref{sec:experiments}, we test our methods with real-world image databases,
and conclude the paper in Section~\ref{sec:conclusion}.

\section{Zero-shot learning with rankings}\label{sec:zero-shot learning}

{\bf Notations}. 
Let $\R$ denote the set of all rankings on $C$ items/classes,
and $\pi=[\pi(1),...,\pi(C)] \in \R$ denote a ranking:
$\pi(i)$ is the position of item $i$ and $\pi^{\text{-1}}(j)$ is the item
number whose position is $j$. 
We write $i \prec j$ (`$i$ precedes $j$') when $\pi(i) < \pi(j)$ (`item $i$ is ranked 
higher than item $j$'.)
A top-K ranking is a straightforward generalization of a ranking, in which the order of
only the first K items $\pi^{\text{-1}}(1), ... , \pi^{\text{-1}}(K)$ matter
and the order of the remaining $C-K$ items are ignored. 
With an abuse of notation, we use $\pi$ and $\R$ as a top-K ranking and the 
as the set of all top-K rankings as well, since a full ranking is a special case 
($K=C$.) 
A partial order is a further generalization of a ranking and a top-K ranking.
In a (full) ranking, a pair of items $(i,j),\;i\neq j$ has to satisfy either $i \prec j$ 
or $j \prec i$, whereas it can be neither in a partial order.
In addition, a partial order has to satisfy the transitivity: 
for any triple $(i,j,k)$, $i \prec j$ and $j\prec k$ implies $i \prec k$.
Item positions $\pi(\cdot)$ are in general undefined for a partial order.


\subsection{Deterministic approach}\label{sec:DR}
A simple deterministic approach to zero-shot learning using semantic rankings was
already outlined in Introduction. 
In one-vs-one setting, $\frac{C(C-1)}{2}$ pre-trained classifiers assign a ranking
$\pi(x)$.
In one-vs-rest, $C$ binary classifiers assign $C$ real-valued scores to 
a test point according to the point's distances to $C$ decision hyperplanes. 
The scores can be sorted to provide a ranking $\pi(x)$. 
Given this ranking $\pi(x)$ of a test sample $x$, and prior knowledge of semantic
rankings $\{\pi_z\;|\; z\in \Z\}$ of test-domain classes $\Z$, we predict
\begin{equation}\label{eq:deterministic ranking}
z(x) = \arg\min_{z \in \Z} \; d(\pi(x),\;\pi_z),  
\end{equation}
where $d(\cdot,\cdot)$ is a distance between two rankings. 
For example, let $\Z=\{{\it cat},\;{\it dog}\}$ whose semantic rankings
are (\ref{eq:ranking of cat}) and (\ref{eq:ranking of truck}), respectively.
If an unseen image $x$ has classification scores in the order
$\pi(x) = [{\it dog}, {\it deer}, \cdots, {\it ship}, {\it airplane}],$
so that $d(\pi(x),\pi_{\mathrm{\it cat}}) < d(\pi(x),\pi_{\mathrm{\it truck}})$
for some $d(\cdot,\cdot)$, then we classify $x$ as a {\it cat} rather than a {\it truck}.
We use the Kendall's ranking distance which is the number of mismatching orders:
\begin{equation}\label{eq:Kendall's distance}
d(\pi_1,\pi_2) =  |\{(i,j)\;|\; \pi_1(i) > \pi_1(j)\; \wedge\; \pi_2(i) < \pi_2(j) \}|.
\end{equation} 
Sometimes it may make more sense to compare only the closest items than to compare all the items.
The top-K version of the Kendall's distance was proposed in \cite{Critchlow:1985:Book},
which can be computed as follows.
Let $A$, $B$, and $D$ be the sets
\begin{eqnarray*}
A &=& \{i \in 1,...,C\;|\;\pi_1(i)\leq K,\; \pi_2(i)\leq K\}\\
B &=& \{i \in 1,...,C\;|\;\pi_1(i)\leq K,\; \pi_2(i)> K\}\\
D &=& \{i \in 1,...,C\;|\;\pi_1(i)> K,\; \pi_2(i)> K\}.
\end{eqnarray*}
Then the Kendall's top-K distance can be computed by
\begin{eqnarray}
d_K(\pi_1,\pi_2) &=& |\{(i,j) \in A\times A| \pi_1(i)\mathrm{<}\pi_1(j),\pi_2(i)\mathrm{>}\pi_2(j)\}| \nonumber\\
&\mathrm{+}& |B|(C\mathrm{+}K\mathrm{-}\frac{|B|\mathrm{-}1}{2})\mathrm{-}\sum_{i\in B} \pi_1(i)\mathrm{-} \sum_{i\in D} \pi_2(i).\label{eq:Kendall's top-K distance}
\end{eqnarray}
Zero-shot classification using the rule (\ref{eq:deterministic ranking}) will be called 
deterministic ranking-based (DR) method.

\subsection{Probabilistic approach}

We can further refine ranking-based algorithms by considering
a probabilistic approach. There are several causes of uncertainty in ranking-based
representation. First, classifier outputs for a test-domain sample $x$ can 
have low confidence, since the classifiers are trained only with training-domain samples. 
Second, prior knowledge of semantic rankings from multiple semantic sources
may not be unanimous.
Third, feature and semantic similarities do not always coincide.
For these reasons, we consider probability models of (top-K) rankings.
We discuss three models: the Mallows \cite{Mallows:1957:Biometrika}, 
the Plackett-Luce \cite{Plackett:1975:AS,Luce:1959:Book,Marden:1995:Book},
and the Babington-Smith \cite{Joe:1993:Book,Smith:1950:JSTOR}, which we will introduce
where they are needed (see \cite{Critchlow:1985:Book,Critchlow:1991:JMP,Marden:1995:Book}
for more reviews.) 

In our probabilistic zero-shot learning approach, we assume the following dependence:
\begin{equation}
    \xymatrix{
	    *+[Fo]{x} \ar[r] & *+[Fo]{\pi} \ar[r] & *+[Fo]{z},
	    }
\end{equation}
that is, the label $z$ of a sample $x$ is dependent only on the predicted ranking $\pi$,
which in turn is dependent only on the sample $x$.
The probability of a test-domain label given the sample $P(z|x)$ is obtained by
marginalizing out the latent ranking variable $\pi$:
\begin{equation}\label{eq:probabilistic zero-shot}
P(z|x) = \sum_{\pi \in \R} P(\pi|x) P(z|\pi) =
\sum_{\pi \in \R} P(\pi|x) \frac{P(z)P(\pi|z)}{\sum_z P(z)P(\pi|z)},
\end{equation}
and the final prediction of the label $z$ for a test sample is made by
$\arg\max_{z}\;P(z|x)$.

There are two terms in (\ref{eq:probabilistic zero-shot}): 
a probabilistic ranker $P(\pi|x)$ and a prior for semantic ranking $P(\pi|z)$.
First, we describe the prior for semantic ranking $P(\pi|z)$ learned from
one or more semantic sources (e.g. different corpora or
crowd-sourcing) in Section~\ref{sec:semantic ranking prior}.
Second, we describe probabilistic rankers $P(\pi|x)$ based on standard classifiers 
trained with training-domain data in Section~\ref{sec:probabilistic ranker}.
The final zero-shot classifier for unseen samples bringing these two learned components
is described in Section~\ref{sec:zero-shot prediction}.

\subsection{Prior for semantic ranking}\label{sec:semantic ranking prior}

We encode the semantic similarity between training- and test-domain classes
by probabilistic ranking models of training-domain classes $P(\pi|z)$
for each test-domain class $z$.
To learn $P(\pi|z)$, we use multiple instances of rankings for each test-domain class.
These rankings can come from multiple linguistic corpora
or by human-rated rankings directly. Below we outline three popular models of rankings
-- the Plackett-Luce, the Mallows, and the Babington-Smith models.

{\bf Plackett-Luce}. 
The Plackett-Luce model for the probability of observing a top-K ranking $\pi$ is 
\begin{equation}\label{eq:PL top-K}
P(\pi;v) = \prod_{i=1}^K \frac{v_{\pi^\text{-1}(i)}}{\sum_{j=i}^C v_{\pi^\text{-1}(j)}}.
\end{equation}
The non-negative parameters $v_1,...,v_C$ indicate the relative chances of being ranked
higher than the rest of the items, and are invariant under constant scaling of $v$'s. 
One interpretation of the generative procedure of the Plackett-Luce model
is the Vase interpretation from \cite{Silberberg:1980:Book}.
Suppose there is a vase with infinite number of balls marked 1 to $C$,
whose numbers are proportional to $v$'s.
At the first stage, a ball is drawn and is recorded as $\pi^\text{-1}(1)$.
At the second stage, another ball is drawn and is recorded as $\pi^\text{-1}(2)$
unless the ball is already selected before ($\pi^\text{-1}(1)$), in which case the drawing
is tried again. 
The procedure is continued until $K$ distinct balls are drawn and recorded.
This generative probability is captured by (\ref{eq:PL top-K}).

Given $N$ samples (=semantic sources) of rankings $\pi_1,...,\pi_N$ for a class,
the parameters can be estimated by MLE.
The log-likelihood of (\ref{eq:PL top-K}) is
\begin{equation}\label{eq:log-likelihood of PL}
L(v) = \sum_{n=1}^N [\sum_{i=1}^K \log (v_{\pi^\text{-1}_n(i)}) 
- \log(\sum_{j=i}^C v_{\pi^\text{-1}_n(j)})],
\end{equation}
with possibly an additional regularization term $\eta \sum v_i^2$.
Hunter \cite{Hunter:2004:AS} proposed an iterative method of estimation using 
the Minorization-Maximization procedure which generalizes the Expectation-Maximization
procedure and converges to a global maximum solution
under a certain condition on the data. 
From our experience, simple gradient-based or Newton-Raphson methods seem to work well
with an appropriate choice of the regularization parameter.

{\bf Mallows}. 
The Mallows model \cite{Mallows:1957:Biometrika} for full rankings is defined as   
$P(\pi;\pi_0,\lambda) \propto e^{-\lambda\;d(\pi,\pi_0)}$,
where $\pi_0$ is the mode, $\lambda$ is the spread parameter,
and $d(\cdot,\cdot)$ is the Kendall's distance between two rankings. 
It may be viewed as a discrete analog of the Gaussian distribution for ranking.
The distance can further be written as
$d(\pi,\pi_0) = d(\pi\pi_0^\text{-1},e) = \sum_{j=1}^{C-1} V_j(\pi\pi_0^\text{-1}),$
where $e$ is the identity ranking and the $V_j$'s are defined as
\begin{equation}\label{eq:V}
V_j(\sigma) = \sum_{i>j} I[\sigma^\text{-1}(i) < \sigma^\text{-1}(j)].
\end{equation}

Fligner et al.~\cite{Fligner:1986:JSTOR} proposed the Mallows model for 
top-K lists by marginalizing the Mallows model:
\begin{equation}\label{eq:Mallows top-K}
P(\pi^{};\pi_0,\lambda) = \frac{1}{\phi^{}(\lambda)} e^{-\lambda \sum_{j=1}^K  V_j(\pi^{}\pi_0^\text{-1})},
\end{equation}
where the $V_j$'s are defined in (\ref{eq:V}) 
and $\phi^{}$ is the normalization constant which can be computed 
in closed form: 
$\phi^{}(\lambda) = \prod_{j=1}^K (1-e^{-(C-j+1)\lambda})/(1-e^{-\lambda}).$

Given $N$ samples of rankings $\pi_1,\pi_2,...,\pi_N$, the parameters of the Mallows model
for total rankings can also be found by MLE~\cite{Feigin:1978:JSTOR}.
When the mode $\pi_0$ is known, the MLE of the spread parameter $\lambda$ can be found 
by convex optimization, 
owing to the fact that the log-likelihood is a concave function of $\lambda$.
The MLE of the centroid $\pi_0$ is the maximum of $\sum_i \log P(\pi_i;\pi_0,\lambda)$
and is equivalent to 
\begin{equation}\label{eq:kemeny optimal}
\arg\min_{\pi_0} \sum_i d(\pi_i,\pi_0).
\end{equation}
The minimization (\ref{eq:kemeny optimal}) is also known as the Kemeny optimal consensus or 
aggregation problem \cite{Kemeny:1959}
and is known to be NP-hard \cite{Bartholdi:1989}. 
However, there are known heuristic methods such as sequential transposition of adjacent
items \cite{Critchlow:1985:Book} or other admissible heuristics \cite{Melia:2007:UAI}.
We use the former method. Starting from the average ranking as the initial value of
$\pi_0$, and we search adjacent items $\pi^\text{-1}(i)$ and $\pi^\text{-1}(i+1)$
whose swapping lowers the sum of distances the most. 
We stop if there is no such item or if the maximum number of iteration (1000 in our case)
is exceeded.
The MLE with (\ref{eq:Mallows top-K}) can be solved by using $\sum_{j=1}^K V_j(\pi)$
 in place of $d$.

{\bf Babington-Smith}. 
The Babington-Smith model \cite{Joe:1993:Book,Smith:1950:JSTOR} is another probabilistic
ranking model based on pairwise comparisons.
Given two items $i$ and $j$, let $\alpha_{ij}$ be the probability that item $i$ is 
ranked higher than item $j$.
Given these preferences $\{\alpha_{ij}\}$, the probability of a ranking $\pi$ is\\
$P(\pi;\alpha) \propto \prod_{i < j} \alpha_{ij}^{I[\pi(i) < \pi(j)]} (1-\alpha_{ij})^{1-I[\pi(i) < \pi(j)]}.$
After introducing new parameters $v_{ij} = \alpha_{ij}/\alpha_{ji}$~\cite{Joe:1993:Book},
the probability of a top-K ranking can be written as \footnote{We presents a slightly modified form.}
\begin{equation}\label{eq:BS top-K}
P(\pi;v) = \frac{1}{\psi(v)}{\prod_{i=1}^K \prod_{j=i+1}^C v_{\pi^\text{-1}(i) \pi^\text{-1}(j)}}
\end{equation}
The Babington-Smith model is similar to the Plackett-Luce model in that the probability
is the product of $v$'s. The larger $v_{ij}$ is, the more likely it is that item $i$ is
ranked higher than item $j$.
However, unlike the Plackett-Luce model, the normalizing constant $\psi(v)$ does not
have a known closed form. 
We do not use it for modeling the semantic prior, but use it for
probabilistic ranker in the next section.

\subsection{Probabilistic ranker from classifiers}\label{sec:probabilistic ranker}

The probabilistic ranker $P(\pi|x)$ takes a sample $x$ as input and 
probabilistically ranks the similarity of $x$ to training-domain classes $\Y$.
We propose to build rankers from standard settings of multiclass classifiers:
one-vs-rest, one-vs-one, or multiclass-loss as in \cite{Crammer:2002:JMLR}.
Any classifier that output a real-valued confidence or score can be used for this purpose.  

{\bf One-vs-rest binary classifiers}. In this setting, there will be $C$ such scores
$f_1(x),....,f_C(x)$ for each training-domain class. 
We relate the real-valued scores $\{f_i\}$ and the non-negative parameters $\{v_i\}$ of 
the Plackett-Luce model (\ref{eq:PL top-K}) by setting $v_i = e^{f_i(x)}$, 
to get 
\begin{equation}\label{eq:PL ranker}
P(\pi|x) = \prod_{i=1}^K \frac{e^{f_{\pi^\text{-1}(i)}(x)}}
{\sum_{j=i}^C {e^{f_{\pi^\text{-1}(j)}(x)}}}.
\end{equation}
Instead of producing a single ranking $\pi$ as in the deterministic approach
(\ref{eq:deterministic ranking}), this ranker evaluates the probability of any
ranking given $x$ taking into account the confidence ($\{f_i(x)\}$) of
$C$ one-vs-rest classifiers.

{\bf One-vs-one binary classifiers}. 
In this setting, there will be $C(C-1)/2$ scores $f_{1,2}(x),....,f_{C-1,C}(x)$
for each pair of training-domain classes.
We related these scores to the $C(C-1)/2$ parameters $\{v_{ij}\}$ of 
the Babington-Smith model (\ref{eq:BS top-K}) by $v_{ij} = e^{f_{ij}(x)}$, 
to get
\begin{equation}\label{eq:BS ranker}
P(\pi|x) \propto 
\prod_{i=1}^K \prod_{j=i+1}^C e^{f_{\pi^\text{-1}(i),\pi^\text{-1}(j)}(x)}.
\end{equation}
Similar to (\ref{eq:PL ranker}), this ranker evaluates the probability of any ranking 
$\pi$ given $x$ taking into account the confidence ($\{f_{ij}(x)\}$) of
$C(C-1)/2$ one-vs-one classifiers.
Note that if the pre-trained classifiers are linear, that is, $f_{ij}(x) = w_{ij}'x$, 
then this ranker is quite similar to (\ref{eq:PL ranker}), since
$P(\pi|x) \propto \prod_{i=1}^K \prod_{j=i+1}^C e^{w_{\pi^\text{-1}(i),\pi^\text{-1}(j)}(x)} 
= \prod_{i=1}^K e^{\hat{w}_{\pi^\text{-1}(i)}'x}$,
with $\hat{w}_{\pi^\text{-1}(i)}$ defined as 
$\sum_{j=i+1}^C w_{\pi^\text{-1}(i),\pi^\text{-1}(j)}$. 
However, it has a different normalization term from (\ref{eq:PL ranker}).

{\bf Multiclass-loss classifiers}. 
Other types of classifiers can be accommodated. When the pre-trained
classifiers are multinomial logistic regression (=softmax) or 
SVMs with a multiclass loss \cite{Crammer:2002:JMLR}, 
we again have $C$ scores $f_1(x),....,f_C(x)$ computed from $C$
parameter vectors $w_1,...,w_C$.
Similar to the one-vs-rest case, we can use the relation $v_i = e^{f_i(x)}=e^{w_i'x}$
with the Plackett-Luce model which gives us the same ranker as (\ref{eq:PL ranker}).
Note that if the original classifier is a multinomial logistic regression, the
(\ref{eq:PL ranker}) is in fact a direct generalization of logistic regression for
$K=1$, which is also observed in \cite{Cheng:2010:ICML}.
In this case, the trained parameters $\{w_i\}$ coincide with the optimal maximum
likelihood parameters for (\ref{eq:PL ranker}) trained with top-1 rankings which are
ground truth labels of the training domain.

To summarize, there exist natural interpretations of the Plackett-Luce and
the Babington-Smith models that allow us to relate classification scores to their
parameters and use them to produce posterior probability $P(\pi|x)$ of rankings
without any further training\footnote{It is not immediately clear whether the Mallows
model can be adapted in this setting and is left for future work.}.

\subsection{Zero-shot prediction}\label{sec:zero-shot prediction}

The probabilistic rankers $P(\pi|x)$ constructed from pre-trained classifiers
and the priors for semantic rankings $P(\pi|z)$ learned from semantic sources
are plugged into (\ref{eq:probabilistic zero-shot}) 
\[
P(z|x) = \sum_{\pi \in \R} P(\pi|x) P(z|\pi) =
\sum_{\pi \in \R} P(\pi|x) \frac{P(z)P(\pi|z)}{\sum_z P(z)P(\pi|z)},
\]
and the final prediction of the label $z$ for a test sample $x$ is made by
$\arg\max_{z}\;P(z|x)$.
The sum over (top-K) rankings $\sum_{\pi \in \R}$ cannot be computed analytically for
either of the Plackett-Luce and the Mallows models 
and requires approximations, e.g., by Markov chain Monte Carlo (MCMC) sampling.
Alternatively, we use $P(\pi|z) = I[\pi = \pi_0^z]$ and a uniform prior $P(z)$,
somewhat similar to \cite{Rohrbach:2010:CVPR}. 
In our preliminary experiments, MCMC-based summation showed inferior results to
this simple version and therefore will be omitted from the report.
The final zero-shot classifier is the MAP classifier 
\begin{equation}\label{eq:final probabilistic zero-shot 1}
\arg\max_z\; \prod_{i=1}^K \frac{e^{f_{(\pi_0^z)^\text{-1}(i)}(x)}}{\sum_{j=i}^C e^{f_{(\pi_0^z)^\text{-1}(j)}(x)}},
\end{equation}
for pre-trained one-vs-rest/multiclass-loss classifiers,
\begin{equation}\label{eq:final probabilistic zero-shot 2}
\mathrm{and}\;\;\;\arg\max_z \prod_{i=1}^K \prod_{j=i+1}^C e^{f_{(\pi_0^z)^\text{-1}(i),(\pi_0^z)^\text{-1}(j)}(x)},
\end{equation}
for pre-trained one-vs-one classifiers.
We summarize the overall training and testing procedures below.

\fbox{%
\begin{minipage}{1\linewidth}
\small
\begin{description}
\item[Training Step 1.] Obtain pre-trained classifiers
	\begin{itemize}
	\item Input: training-domain sample and label pairs $\{(x_1,y_1),...,(x_N,y_N)\}$, 
	regularization hyperparameter 
	\item Output: score functions $f_1,...,f_C$ or $f_{1,2},...,f_{C-1,C}$ 
	\item Method: one-vs-rest/one-vs-one/multiclass with any classifier
	\end{itemize}
\item[Training Step 2.] Learn priors for semantic rankings
	\begin{itemize}
	\item Input: ranking and test-domain label pairs $\{(\pi_1,z_1),...,(\pi_M,z_M)\}$
	 collected from corpora or crowdsourcing
	\item Output: consensus rankings $\pi_0^z$ for each test-domain class $z=1,...,L$
	from either the Plackett-Luce model (\ref{eq:PL top-K}) or the Mallows (\ref{eq:Mallows top-K})
	\item Method: MLE of (\ref{eq:log-likelihood of PL}) by BFGS or 
	MLE of (\ref{eq:kemeny optimal}) by sequential transposition
	\end{itemize}
\item[Testing.] Zero-shot classification
	\begin{itemize}
	\item Input: data $x$, parameter $K$ for top-K list size
	\item Output: prediction of test-domain label $z$
	\item Method: MAP estimation (\ref{eq:final probabilistic zero-shot 1}) or (\ref{eq:final probabilistic zero-shot 2}), 
	using $f(x)$'s from Training Step 1 and $\pi_0^z$'s from Training Step 2
	\end{itemize}
\end{description}
\end{minipage}
}

\section{Related work}\label{sec:related work}

There are two major approaches to zero-shot learning explored
in the literature: attribute-based and similarity-based. 
In attribute-based knowledge transfer (e.g., \cite{Palatucci:2009:NIPS,Lampert:2009:CVPR,Akata:2013:CVPR}), 
the classes from training and test domains 
are assumed to be distinguishable by a common list of attributes.
Attribute-based approaches often show excellent empirical
performance~\cite{Palatucci:2009:NIPS,Rohrbach:2010:CVPR}.
However, designing the attributes that are discriminative, common to multiple classes,
and correlated with the original feature at the same time, can be a non-trivial task that
typically requires human supervision. Similar arguments can be found in 
\cite{Rohrbach:2010:CVPR} or \cite{Mensink:2014}.

By contrast, similarity-based zero-shot approaches use semantic similarity between
training-domain classes $\Y$ and test-domain classes $\Z$ directly. 
The advantage of this approach is that similarity information can be mined automatically
from the web, linguistic corpora or other sources. 
Similarity information has been used to build a probabilistic zero-shot classifier
called direct similarity-based method (DS) \cite{Rohrbach:2010:CVPR,Rohrbach:2011:CVPR},
which parallels the attribute-based approach from \cite{Lampert:2009:CVPR}.
Direct similarity-based method also uses classification scores and probabilistic inference
as ours, but it uses numerical similarity instead of non-metric ranking presentation
in our method. 
More recently, similarity-based approaches using semantic embedding have been proposed
~\cite{Frome:2013:NIPS,Socher:2013:NIPS}. In these algorithms, training and test domain
objects are simultaneously embedded into a semantic space using multilayer 
neural networks. While these two methods produce interesting results, they use
specific metric similarity models and require retraining when the semantic model changes,
unlike our method. 
Mensink et al.~use a linear combination of pre-trained classifiers for
classifying unseen data \cite{Mensink:2014}. 
They use co-occurrence statistics as semantic information, 
whereas we do not assume a specific type of similarity information. 
Lastly, our method provides a means to aggregate multiple semantic sources that
has not been addressed in the literature. 

\section{Experiments}\label{sec:experiments}

\subsection{Datasets}
We use two datasets 
1) Animals with Attributes dataset \cite{Lampert:2009:CVPR} and
2) CIFAR-100/10 \cite{Torralba:2008:TPAMI} collected by \cite{Krizhevsky:2009:TR}.
Semantic similarity is obtained from WordNet distance, 
web searches~\cite{Rohrbach:2010:CVPR}, word2vec~\cite{Mikolov:2013}, GloVe~\cite{Pennington:2014}, and from Amazon Mechanical Turk.
Table~\ref{tbl:datasets} summarizes the characteristics of the datasets and the types of 
available semantic information used in the experiments. 
More details on data processing are provided in Appendix.

\begin{table*}[tb]
\caption{Datasets used in the experiments.}
\label{tbl:datasets}
\begin{center}
\begin{small}
\begin{tabular}{|l|p{2.00in}|p{2.00in}|}
\hline
& Animals & CIFAR \\
\hline
Feature dimension & 8941 & 4000 \\
Number of training/validation/test samples & 21847 / 2448 / 6180 & 50000 / 50000 / 10000 \\
Number of training/test classes & 40 / 10 & 100 / 10 \\
\hline
Linguistic sources &  
WordNet, Wikipedia, Yahoo, YahooImage, Flicker
& WordNet, word2vec~\cite{Mikolov:2013}, GloVe~\cite{Pennington:2014}\\
\hline
Number of surveys from crowd-sourcing & 500 & 500 \\
\hline
\end{tabular}
\end{small}
\end{center}
\end{table*}

\subsection{Methods}

We perform comprehensive tests of the probabilistic ranking-based (PR) zero-shot
model under 1) three learning settings (one-vs-rest, one-vs-one, multiclass), 
2) two types of semantic sources (linguistic, crowd-sourcing), and 3) different prior models for semantic rankings (the Plackett-Luce and the Mallows models). 
We compare probabilistic ranking-based method (PR, Sec.~\ref{sec:zero-shot prediction}) 
to deterministic ranking-based method (DR, Sec.~\ref{sec:DR})and direct similarity-based method
(DS, \cite{Rohrbach:2010:CVPR})
which is the closest state-of-the-art to our methods that uses classifier scores.
We also refer to other results in the literature for comparison.

Regularization parameters for classifiers are determined from
the validation set and partially manually to avoid exhaustive cross-validation.
We test with different hyperameters $K$ (in top-K list) and report 
the results with $K=4$. 
For one-vs-rest and one-vs-one, we trained SVMs followed by Platt's probabilistic scaling
~\cite{Platt:1999}. For multiclass, we used multinomial logistic regression.

\subsection{Result 1 -- Discriminability of semantic rankings}

\begin{figure}[t]
\begin{center}
\includegraphics[width=1\linewidth]{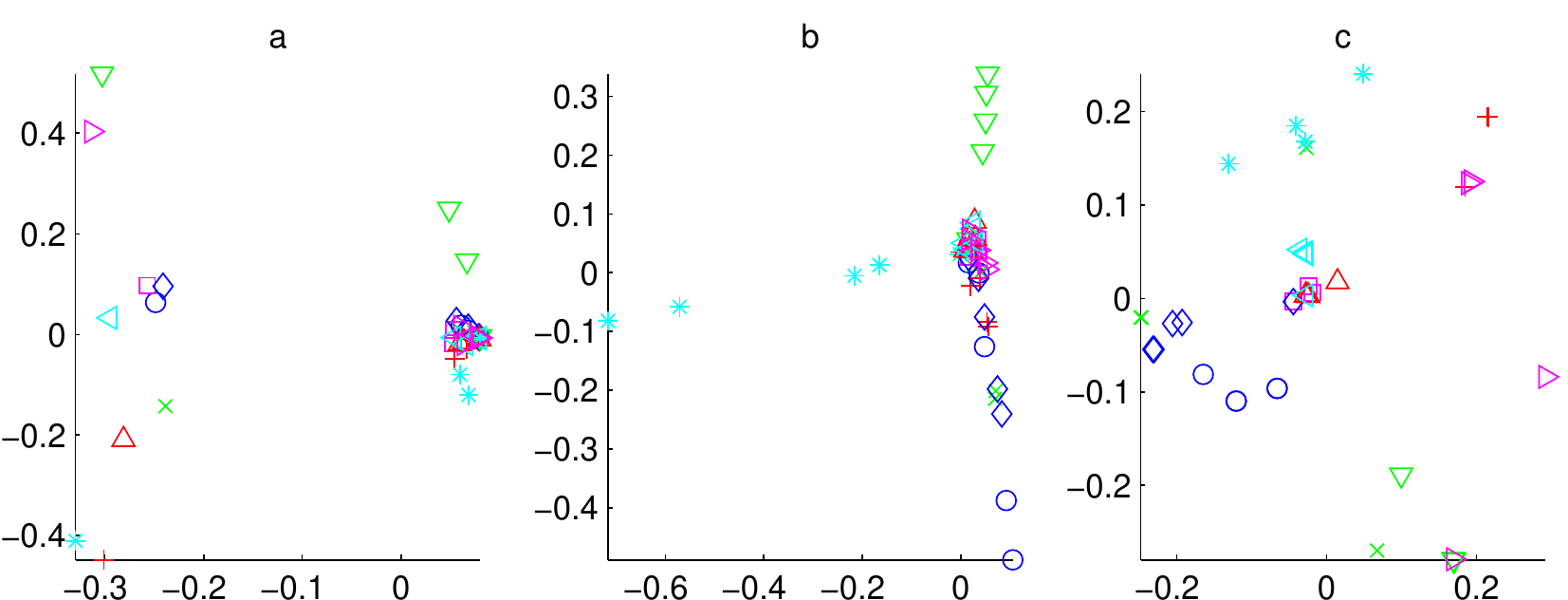}
\end{center}
\caption{Multidimensional scaling of similarity data for Animals, using 
Euclidean distance (a), Euclidean distance with normalization, and 
Kendall's distance (c)($K=2$).
The 10 test-domain classes are plotted with different symbols and colors.
}
\label{fig:mds}
\end{figure}

\begin{table*}[tb]
\caption{Zero-shot classification accuracy of Direct Similarity (DS), Deterministic
Ranking (DR), and Probabilistic Ranking-based (RP).
Each method is tested with different semantic source and classifier types.
{\it Indiv}: averaged accuracy using individual semantic similarities, {\it Arithm}: accuracy using 
arithmetic mean of similarities, {\it Geom}: accuracy using geometric mean of similarities.
The best result for each method is highlighted in boldface.
}
\label{tbl:comparison}
\begin{center}
\begin{small}
\begin{tabular}{c|c|c|c||c|c|c||c|c|c|c|}
\multicolumn{11}{c}{{\bf Animals dataset}} \\
\cline{2-11}
 & \multicolumn{3}{c||}{Direct Similarity (DS)} & \multicolumn{3}{|c||}{Deterministic Ranking (DR)} & \multicolumn{4}{|c|}{Probabilistic Ranking (PR)}\\
\cline{2-11}
\cline{2-11}
& \multicolumn{3}{|c||}{Linguistic sources} & \multicolumn{3}{|c||}{Linguistic sources} & \multicolumn{2}{|c|}{Linguistic sources} & \multicolumn{2}{|c|}{Crowd-source}\\
\cline{2-11}
& Indiv & Arithm & Geom & Indiv & Arithm & Geom & P.-L & Mallows & P.-L & Mallows \\
\hline
\multicolumn{1}{|c|}{one-vs-rest} & 0.320 & 0.334 & {\bf 0.354} & 0.329 & 0.330 & 0.347 & 0.320 & 0.312 & 0.351 & 0.351\\
\hline
\multicolumn{1}{|c|}{one-vs-one} & \multicolumn{3}{c||}{n/a} & 0.341 & 0.343 & {\bf 0.359} & 0.358 & 0.320 & 0.374 & 0.374\\
\hline
\multicolumn{1}{|c|}{multiclass} & \multicolumn{3}{c||}{n/a} & 0.331 & 0.345 & 0.355 & {\bf 0.370} & 0.345 & {\bf 0.395} & 0.392 \\
\hline
\multicolumn{11}{c}{}\\
\multicolumn{11}{c}{{\bf CIFAR dataset}} \\
\cline{2-11}
 & \multicolumn{3}{c||}{Direct Similarity (DS)} & \multicolumn{3}{|c||}{Deterministic Ranking (DR)} & \multicolumn{4}{|c|}{Probabilistic Ranking (PR)}\\
\cline{2-11}
\cline{2-11}
& \multicolumn{3}{|c||}{Linguistic sources} & \multicolumn{3}{|c||}{Linguistic sources} & \multicolumn{2}{|c|}{Linguistic sources} & \multicolumn{2}{|c|}{Crowd-source}\\
\cline{2-11}
& Indiv & Arithm & Geom & Indiv & Arithm & Geom & P.-L & Mallows & P.-L & Mallows \\
\hline
\multicolumn{1}{|c|}{one-vs-rest} & 0.273 & 0.300 & {\bf 0.316} & 0.224 & 0.258 & 0.260 & 0.314 & 0.288 & 0.258 & 0.282 \\
\hline
\multicolumn{1}{|c|}{one-vs-one} & \multicolumn{3}{c||}{n/a} & 0.244 & {\bf 0.281} & 0.278 & 0.335 & 0.297 & 0.244 & 0.261 \\
\hline
\multicolumn{1}{|c|}{multiclass} & \multicolumn{3}{c||}{n/a} & 0.251 & 0.272 & 0.276 & {\bf 0.339} & 0.320 & 0.260 & {\bf 0.292}\\
\hline
\end{tabular}
\end{small}
\end{center}
\end{table*}

We first compare the discriminability of classes with ranking vs numerical 
representations of similarity without using image data.
Using all five linguistic sources for Animals, we compute pairwise distances
of 5$\times$10=50 similarity vectors. Two types of distances are computed --
the Euclidean distance of numerical similarity, with or without $l_1$ normalization, 
and the Hausdorff distance for top-K lists using 
the Kendall's ranking distance (\ref{eq:Kendall's top-K distance}).
Note that the rankings are obtained by sorting the numerically similarity.
For these different representations, 
the average accuracy of leave-one-out 1-Nearest Neighbor classification was
0.44 (Euclidean), 0.62 (Euclidean with normalization), 
0.72 (Kendall's, $K$=2), 0.70 (Kendall's, $K$=5), and 0.64 (Kendall's, $K$=10),
which shows that the ranking distances are better than the Euclidean distances
for discriminating test-domain classes when there are multiple heterogeneous sources.
Figure~\ref{fig:mds} shows the embeddings from classical Multidimensional Scaling (MDS)
using these distances. It shows qualitative differences of numerical similarity 
(a and b) and ranking (c). 
The embedding of rankings has better between-class separation and
within-class clustering than the embeddings of numerical similarity,
suggesting that the non-metric order information is more consistent than numerical
similarity across different sources.

We also compute leave-one-out accuracy of Bayesian classification with 
the rankings collected directly from crowd-sourcing for Animals and CIFAR datasets.
Out of 500 rankings, one ranking is held out and the 499 remaining rankings are
used to build 10 semantic ranking probabilities $P(\pi|z)$ using both
the Mallows and the Plackett-Luce models.
Prediction of the class of the held-out rankings is made by the maximum of $P(\pi|z)$
over all 10 classes.
For Animals, the average accuracy was 0.91/0.99/0.99 ($K$=2/5/10) 
using the Mallows model, and 0.79/0.84/0.96 ($K$=2/5/10) using the Plackett-Luce model.
For CIFAR, the average accuracy was 0.73/0.79/0.84 ($K$=2/5/10) 
using the Mallows model, and 0.72/0.77/0.74 ($K$=2/5/10) using the Plackett-Luce model.
These numbers show that the rankings obtained from crowd-sourcing have 
information to discriminate the test-domain classes with up to $0.8\sim 1.0$ accuracy.

\subsection{Result 2 - Comparison of PR, DR, and DS}
We compare probabilistic ranking (PR), deterministic ranking (DR) and direct similarity
(DS) methods for zero-shot classification accuracy. 
All three methods share the same image features and the same linguistic sources
of semantic information (except for the crowd-sourcing for PR), 
but use them in different ways. 
PR uses probabilistic models to combine multiple sources of semantic similarity. 
DR and DS inherently use a single source of semantic similarity, and therefore the multiple
sources have to be combined heuristically. We first normalize individual similarity sources
to be in the range from 0 to 1, and then compute arithmetic and geometric
means over multiple sources. 
Note that the main difference between DR and DS, is that DR uses rankings whereas 
DS uses numeric values.

{\bf DS vs DR}. 
The results are shown in Table~\ref{tbl:comparison}.
For both DR and DS, using averaged semantic similarity ({``Arithm''} and {``Geom''})
is better than using individual similarity ({``Indiv''}), for both Animals and CIFAR datasets.
A plausible interpretation is that the aggregate similarity is more reliable than individual
similarities despite using heuristic methods of aggregation. 
The highest accuracy from DS is 0.354 (for Animals) and 0.316 (for CIFAR), 
whereas the hight accuracy from DR is 0.359 (for Animals) and 0.281 (for CIFAR).
DR performs slightly better than DS in Animals, but worse in CIFAR.
Within DR, accuracy is not affected much by the pre-trained classifier type
(one-vs-rest, one-vs-one, multiclass). 

{\bf PR vs others}. 
Using the same linguistic sources, the highest accuracy from PR is
0.370 (for Animals) and 0.339 (for CIFAR) which are much higher than DS and DR
regardless of whether a single (Indiv) or multiple (Arithm and Geom) sources are used.
This suggests the advantage of using probabilistic models to aggregate multiple
semantic sources.
Within PR results, one-vs-rest and one-vs-one classifiers perform comparably, and
multiclass logistic regression performs the best. 
PR performs even better with crowd-sourced semantic information (0.395)
than with linguistic sources (0.370) in Animals, but the opposite is true in CIFAR, 
probably due to the less reliability of human subject ratings with CIFAR (sorting 100 categories 
correctly compared to 40 in Animals).

In the literature, the accuracy of attributes-based methods with Animals
 ranges from 0.36 to 0.44 (Tables 3 and 4, \cite{Akata:2013:CVPR}), 
compared to 0.395 from our method which do not use attributes.
We remind the reader that finding `good' attributes is itself a non-trivial task.
When both similarity and attributes are mined automatically from corpora, 
similarity-based methods perform much better than attributed-based methods 
(individual average of 0.22 from Table 1, \cite{Rohrbach:2010:CVPR}).

\begin{figure}[t]
\begin{center}
\includegraphics[width=1.0\linewidth]{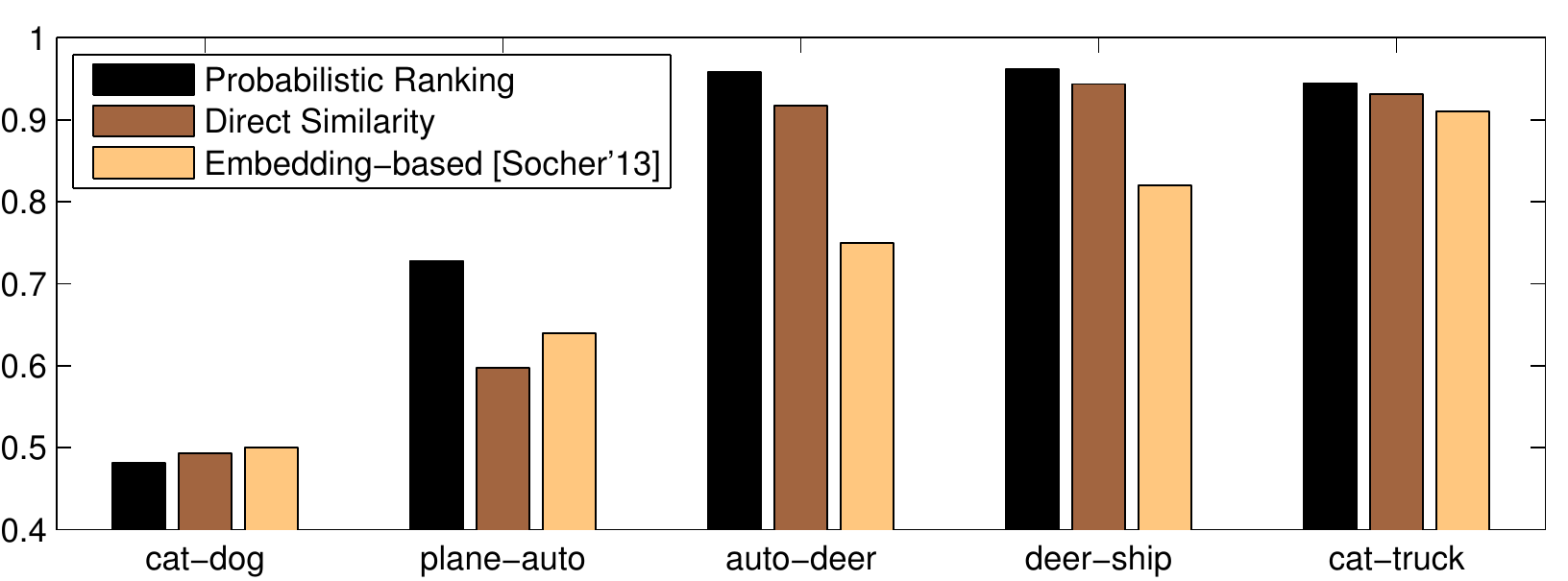}
\end{center}
\caption{Zero-shot classification accuracy for CIFAR.}
\label{fig:twoclass}
\vspace{-0.2in}
\end{figure}

Lastly, Figure~\ref{fig:twoclass} shows two-class classification accuracy 
of PR (PL$+$linguistic sources), DS, and an embedding-based method
on select pairs of classes from CIFAR (Figure~3, \cite{Socher:2013:NIPS}).
Although the numbers may not be directly comparable due to different settings\footnote{Socher et al.~used the rest of classes from CIFAR-10 instead of CIFAR-100 for training, 
and also used different semantic information}, PR performs noticeably better than 
the two state-of-the-arts. In fact, we can distinguish {\it auto} vs {\it deer}, 
{\it deer} vs {\it ship}, or {\it cat} vs {\it truck} with $\sim 95\%$
accuracy, without a single training image of these categories.

\section{Conclusion}\label{sec:conclusion}

In this paper, we propose ranking-based representation of semantic similarity,
as an alternative to metric representation of similarity.
Using rankings, semantic information from multiple sources can be aggregated naturally
to produce a better representation than individual sources. 
Using this representation and probability models of rankings, 
we present new zero-shot classifiers which can be constructed from pretrained
classifiers without retraining, and demonstrate their potential for exploiting
semantic structures of real-world visual objects. 

%

{\small
\bibliographystyle{icml2015}
\bibliography{icml15_jh}
}

\appendix 
\section{Datasets}
The Animals with Attributes dataset (Animals) was collected and processed by \cite{Lampert:2009:CVPR}.
The training domain consists of images of 40 types of animals,
from which 21,847 and 2,448 images were used as training and validation sets.
From each image, 10,942 dimensional features are extracted \cite{Lampert:2009:CVPR}.
The test domain consists of 6,180 images of 10 types of animals which are
non-overlapping with the training-domain classes.
Semantic similarity of the animals are provided by \cite{Rohrbach:2010:CVPR}
\footnote{\url{http://www.d2.mpi-inf.mpg.de/nlp4vision}},
which are computed from five different linguistic sources:
path distance from WordNet, co-occurrence from Wikipedia, Yahoo web search,
Yahoo image Search, and Flickr image search.

The CIFAR-100 and CIFAR-10 are collected by \cite{Krizhevsky:2009:TR}.
The training domain (CIFAR-100) consists of 60,000 images of 100 types of objects 
including animals, plants, household objects, and scenery.
We use 50,000 and 10,000 images from CIFAR-100 as training and validation sets.
The test domain (CIFAR-10) consists of 60,000 images of 10 types of objects similar
to CIFAR-100, without any overlap with the classes from CIFAR-100.
We use 10,000 images as test data.
To compute features, we use a deep-trained neural network
\footnote{\url{https://github.com/jetpacapp/DeepBeliefSDK}}, which is trained from 
ImageNet ILSVRC2010 dataset\footnote{\url{http://www.image-net.org/challenges/LSVRC}}
consisting of 1.2 million images of 1000 categories.
We apply CIFAR-100 and CIFAR-10 training images to the network, and
use the 4096-dimensional output from the last hidden layer of the network as features.
For semantic similarity of CIFAR-100 and CIFAR-10, we compute the WordNet path distance,
and also used word2vec tools~\cite{Mikolov:2013}
\footnote{\url{https://code.google.com/p/word2vec/}}
and GloVe tools~\cite{Pennington:2014}\footnote{\url{http://nlp.stanford.edu/projects/glove/}}.

In addition to using linguistic sources, 
we use Amazon Mechanical Turk to collect word similarity data by crowd-sourcing.
Each participant of the survey is shown a word from the test domain classes,
and is asked to sort 10 words from the training domain according to their perceived similarity
to the given word. The initial order of 10 words is randomized for each survey. 
We pre-select those 10 closest words for each test-domain word, because we found
from preliminary trials that ordering all words (40 for Animal and 100 for CIFAR)
is too demanding and time-consuming for participants.
For Animal, 10 closest words are selected based on the average ranking of the words w.r.t. the
test-domain word from the five linguistic sources. 
For CIFAR, we use the path distance from WordNet.
Fifty surveys are collected for each test-domain class.

\section{Implementation}
The Direct Similarity-based method (DS) \cite{Rohrbach:2010:CVPR} is implemented as follows. 
The probability $P(y_k|x)$ is modeled by a 
one-vs-rest binary SVM classifier followed by the Platt's probabilistic scaling
 \cite{Platt:1999}, trained with training-domain feature and label pairs.
In testing, the probability $P(y_k|x)$ is evaluated for a test image,
and the prediction of the test-domain class is made by MAP estimation using
\begin{equation}\label{eq:rohrbach}
P(z|x) \propto \prod_{k=1}^K \left(\frac{P(y_k|x)}{P(y_k)}\right)^{y_k^z},\;
y^z_k = w^z_{y_k} / \sum_{i=1}^5 w^z_{y_i}
\end{equation}
where $w^z_{y_k}$ is the similarity score. 
The sum above is limited to five most similar training-domain classes.
We have tested different values of the prior $P(y_k)$, which did not have
visible effects on the result.

%
\end{document}